\documentclass[10pt,twocolumn,letterpaper]{article}

\usepackage[pagenumbers]{wacv} 

\usepackage{graphicx}
\usepackage{amsmath}
\usepackage{amssymb}
\usepackage{booktabs}
\usepackage{xcolor}         
\usepackage{multirow}
\usepackage{array, makecell}
\usepackage{colortbl}
\usepackage{threeparttable}
\usepackage{cuted}
\usepackage{etoolbox}
\AfterEndEnvironment{strip}{\leavevmode}

\usepackage[pagebackref,breaklinks,colorlinks]{hyperref}
\usepackage[nameinlink,capitalize]{cleveref}

\usepackage[capitalize]{cleveref}
\crefname{section}{Sec.}{Secs.}
\Crefname{section}{Section}{Sections}
\Crefname{table}{Table}{Tables}
\crefname{table}{Tab.}{Tabs.}

\Crefformat{equation}{#2Eq.~(#1)#3}
\Crefformat{figure}{#2Fig.~#1#3}

\begin{document}

\title{AgileFormer: Spatially Agile Transformer UNet for Medical Image Segmentation}
\vspace{-0.4cm}
\author{Peijie Qiu, Jin Yang, Sayantan Kumar, Soumyendu Sekhar Ghosh, Aristeidis Sotiras\\
Washington University in St. Louis, St. Louis, MO, USA \\
\href{https://github.com/sotiraslab/AgileFormer} {https://github.com/sotiraslab/AgileFormer}
}
\maketitle


\begin{strip}
\begin{minipage}{\textwidth}\centering
\vspace{-0.5cm}
\includegraphics[width=1.0\textwidth]{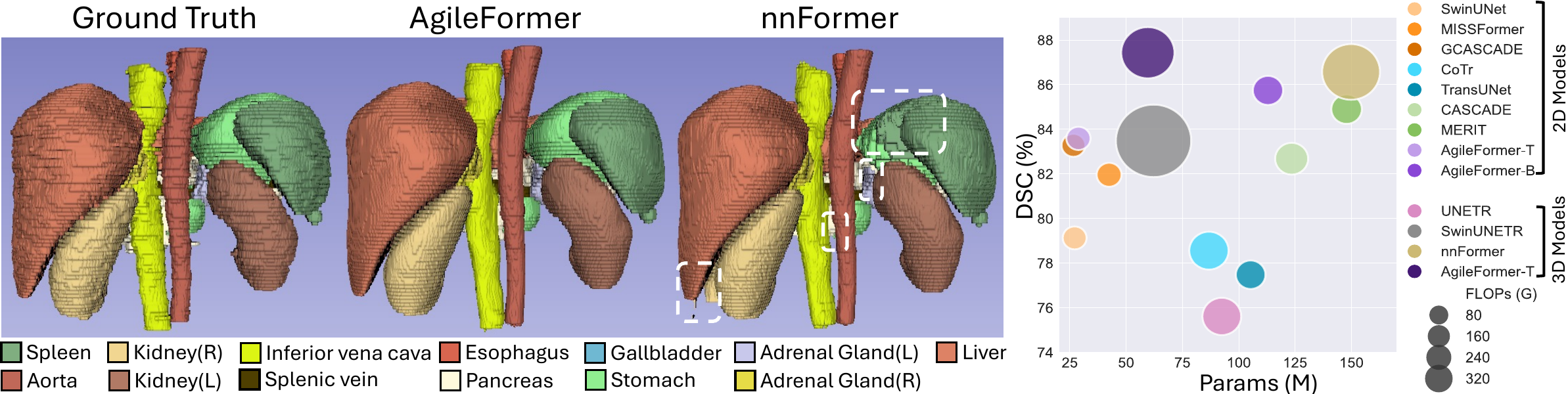}
\captionof{figure}{\textbf{Left}: Qualitative comparison between the proposed AgileFormer and nnFormer~\cite{zhou2021nnformer} on the Synapse multi-organ segmentation task, where each white dashed box marks inaccurate segmented regions. For demonstration purposes, we visualize all 13 organs, but we only report results for 8 of them. We can observe that nnFormer struggled to accurately segment the spleen and stomach,  confusing one for the other. Moreover, it over-segmented the right kidney. This is likely due to the fact that fixed-sizing window attention and patch embedding cannot accurately capture objects with varying sizes and shapes, and hence produce inaccurate feature representations. In contrast,  AgileFormer, which can capture spatially varying representations via deformable patch embedding, spatially dynamic self-attention and multi-scale deformable positional encoding, accurately segmented organs with varying sizes and shapes. \textbf{Right}: Segmentation accuracy (DSC) against model complexity (number of parameters and FLOPs) on the Synapse multi-organ segmentation task. In both 2D and 3D settings, AgileFormer outperformed recent state-of-the-art methods, while having fewer parameters and FLOPs.}
\label{fig:graphic_abstract}
\end{minipage}
\end{strip}

\begin{abstract}
\vspace{-0.5cm}
    In the past decades, deep neural networks, particularly convolutional neural networks, have achieved state-of-the-art performance in a variety of medical image segmentation tasks. Recently, the introduction of vision transformers (ViTs) has significantly altered the landscape of deep segmentation models. The excellent performance and scalability of ViTs have driven a growing focus on their application. However, we argue that the current design of the vision transformer-based UNet (ViT-UNet) segmentation models may not effectively handle the heterogeneous appearance (e.g., varying shapes and sizes) of target objects. In addition, prevailing ViT-UNets focus on enhancing the self-attention building block while neglecting other important components (i.e., patching embedding and positional encoding). To tackle these limitations, we present a structured approach to introduce spatially dynamic components into a ViT-UNet. This enables the model to effectively capture features of target objects with diverse appearances. This is achieved by three main components: \textbf{(i)} deformable patch embedding; \textbf{(ii)} spatially dynamic multi-head attention; \textbf{(iii)} multi-scale deformable positional encoding. These components are integrated into a novel architecture, termed \textbf{AgileFormer}. Experiments in three segmentation tasks using publicly available datasets (Synapse multi-organ, ACDC cardiac, and Decathlon brain tumor datasets) demonstrated the effectiveness of AgileFormer for 2D and 3D segmentation tasks. Remarkably, AgileFormer sets a new state-of-the-art performance with a Dice Score of \textbf{85.74\%} and \textbf{87.43 \%} for 2D and 3D multi-organ segmentation on Synapse without significant computational overhead. 
\end{abstract}


\section{Introduction} \label{sec:intro}
Medical image segmentation tasks are important in modern medicine, as they typically serve as the first step in many image-driven diagnoses and analyses~\cite{chen2021transunet,cao2022swin}. Deep learning based automated segmentation methods have dominated this field due to their high efficiency and state-of-the-art performance. Among them, convolutional neural networks (CNNs)~\cite{ronneberger2015u,oktay2018attention,lee20223d,isensee2018brain,isensee2021nnu} have emerged as the prevalent choice since the introduction of UNet~\cite{ronneberger2015u}. This is due to CNNs' inherent advantages in handling image-driven tasks, such as their ability to capture locality and translation invariance.
However, they struggle with capturing global semantics primarily due to their restricted receptive field. 

In contrast, the recently proposed vision transformer (ViT)~\cite{dosovitskiy2020image} mitigates this problem through a self-attention mechanism that captures dependencies between image patches regardless of their spatial distances. 
The first ViT-based UNet (ViT-UNet) for medical image segmentation combined a ViT encoder with a CNN decoder, and hence was termed TransUNet~\cite{chen2021transunet}. However, TransUNet is buderened by a large number of parameters ($\sim$100M) and considerable computational complexity. This is because it employs standard self-attention, which has quadratic time and memory complexity w.r.t. the input token size. 
SwinUNet~\cite{liu2021swin} addressed this challenge by leveraging window attention~\cite{liu2021swin}, which performs self-attention within small windows in parallel for all image patches. SwinUNet is the first pure ViT-UNet, with self-attention as the main feature extractor for the encoder, decoder, and bottleneck of a UNet. 
However, SwinUNet uses fixed-size windowing, which may limit its ability to capture precise representations for target objects with varying sizes and shapes. 
Additionally, previous ViT-UNets deploy a fixed-size patch partitioning (i.e., patch embedding), which inherently limits their ability to localize objects with varying sizes and shapes (see evidence in~\Cref{fig:graphic_abstract} (\textbf{Left}) and~\Cref{fig:segm_vis} (a)). Because objects are not always perfectly bounded by fixed-size square patches. 

We argue that the inability of standard ViT-UNets to capture spatially heterogeneous representations restricts their scalability and adaptability to various segmentation tasks, especially those involving heterogeneous targets. One notable example is the multi-organ segmentation task, where TransUNet and SwinUNet do not exhibit good scalability when increasing the number of parameters (as evident in \Cref{fig:scalability}). In contrast, models that can capture spatially varying representations (i.e., CoTr~\cite{xie2021cotr} and MERIT~\cite{rahman2024multi})  showed better scalability. However, the building block in the encoder and decoder of CoTr remains convolutional layers with the bottleneck replaced by a Deformable DETER~\cite{zhu2020deformable}. MERIT~\cite{rahman2024multi} only captures multi-scale feature representations by having a multi-resolution input, which may fail to capture feature representation with varying shapes. We kindly direct readers to \texttt{Appendix A} for a more detailed comparison of the proposed method with CoTr and MERIT. Furthermore, the adaptability of ViT-UNet deteriorates when training samples are extremely scarce. This is particularly true in 3D multi-organ segmentation with only 18 training samples, where SwinUNETR~\cite{hatamizadeh2021swin} and nnFormer~\cite{zhou2021nnformer} did not empirically show superior performance compared to CNN-based nnUNet~\cite{isensee2021nnu}. 
We conjecture that the hierarchical feature extraction in CNNs naturally captures structures within objects at multiple scales. However, this is more challenging for fixed-size window attention and patch embedding in SwinUNet and nnFormer.



\noindent \textbf{Contributions:} To address limitations in previous ViT-UNet designs, we propose a spatially agile pure ViT-UNet to capture diverse target objects in a medical image segmentation task. We introduce a structured approach to introduce spatially dynamic components that can capture objects with varying sizes and shapes into a standard ViT-UNet. First, we replaced the standard rigid square patch embedding in ViT-UNet with a novel deformable patch embedding. Second, we adopted spatially dynamic self-attention~\cite{xia2023dat++} as the building block to capture spatially varying features. Third, we proposed a novel multi-scale deformable positional encoding to model irregularly sampled grids in self-attention.
We integrated these dynamic components into a novel ViT-UNet architecture, named \textbf{AgileFormer}.

We validated our method on three medical image segmentation tasks, including multi-organ, cardiac, and brain tumor MRI segmentation.
Extensive experiments demonstrated the effectiveness
of the proposed method for three medical image segmentation tasks. Specifically, AgileFormer outperformed recent state-of-the-art UNet models for both 2D and 3D medical image segmentation across all three segmentation tasks, demonstrating exceptional scalability. Remarkably, AgileFormer set a new state-of-the-art performance for multi-organ segmentation, achieving an average DSC (\%) of 85.74 and 87.43 in 2D and 3D settings with fewer parameters and FLOPs (see~\Cref{fig:graphic_abstract} (\textbf{Right})). 
Importantly, our empirical investigations also revealed critical components (i.e., patch embedding and positional encoding) in designing a ViT-UNet, which was neglected by previous works but has a significant impact on performance.

\section{Related work}
\noindent \textbf{CNN-based segmentation methods:} During the past decade, convolutional neural networks (CNNs) have been considered the defacto standard for medical image segmentation following the emergence of UNet~\cite{ronneberger2015u}. Many follow-up works extend the U-shaped encoder-decoder network by introducing different skip-connection methods, including attention-based skip-connection (e.g., Attn-UNet~\cite{oktay2018attention} and UCTransUNet~\cite{wang2022uctransnet}) and dense skip-connection (e.g., UNet++~\cite{zhou2019unet++} and UNet 3+~\cite{huang2020unet}). Another line of works modified the convolution building block in UNet by introducing residual connection~\cite{zhang2018road,isensee2018brain}, deformable convolution~\cite{li2020pgd}, and large kernel convolution~\cite{li2022lkau,lee20223d,yang2024d}. In addition, methods that leverage multi-scale information for capturing representation at varying sizes have been investigated~\cite{zhao2017pyramid,roth2017hierarchical}. Previous work also explored a generalized UNet framework, termed nnUNet~\cite{isensee2021nnu}, which provides a self-configuring UNet framework for any dataset. Finally, MedNeXt~\cite{roy2023mednext} extended nnUNet by proposing a transformer-like architecture design of nnUNet following the advances in model designs outlined in  ConvNeXt~\cite{liu2022convnet}.

\noindent \textbf{ViT-based segmentation methods:} Since its introduction in 2020, ViT~\cite{dosovitskiy2020image} has significantly shifted the trajectory of the design of UNets. This is because self-attention in ViTs is more advantageous in capturing long-range dependencies between image patches than convolution in CNNs. In particular, early explorations integrated ViTs into traditional CNN-based UNets by replacing the CNN encoder~\cite{chen2021transunet,hatamizadeh2022unetr} and bottleneck~\cite{wang2021transbts,xie2021cotr} with a ViT. There are also hybrid models~\cite{heidari2023hiformer,xie2021cotr,rahman2023medical,rahman2024g}, where the main feature extraction in the encoder/decoder is achieved by both CNN and ViT. 
However, due to the quadratic complexity and the lack of locality in standard self-attention, few pure ViT-based UNets~\cite{cao2022swin,zhou2021nnformer,huang2022missformer,rahman2024multi} have been proposed for medical image segmentation tasks, with most relying on the shifted window attention~\cite{liu2021swin}. 

However, prevailing methods focus on advancing the self-attention building block in a ViT-UNet, while neglecting the other important components (e.g., patch embedding and positional encoding) in building a ViT-UNet. In contrast, our empirical findings revealed that the design of the patch embedding and positional encoding were equally important as the self-attention in enhancing the performance of a ViT-UNet. In addition, none of the prior ViT-UNets for medical image segmentation has explored capturing spatially dynamic feature representation to handle segmentation targets of varying sizes and shapes. In contrast, our Agileformer provides a systematical way to capture spatially dynamic feature representation by a novel deformable patch embedding, spatially dynamic self-attention, and deformable positional encoding. 

\begin{figure}[!t]
    \centering
    \includegraphics[width=1\columnwidth]{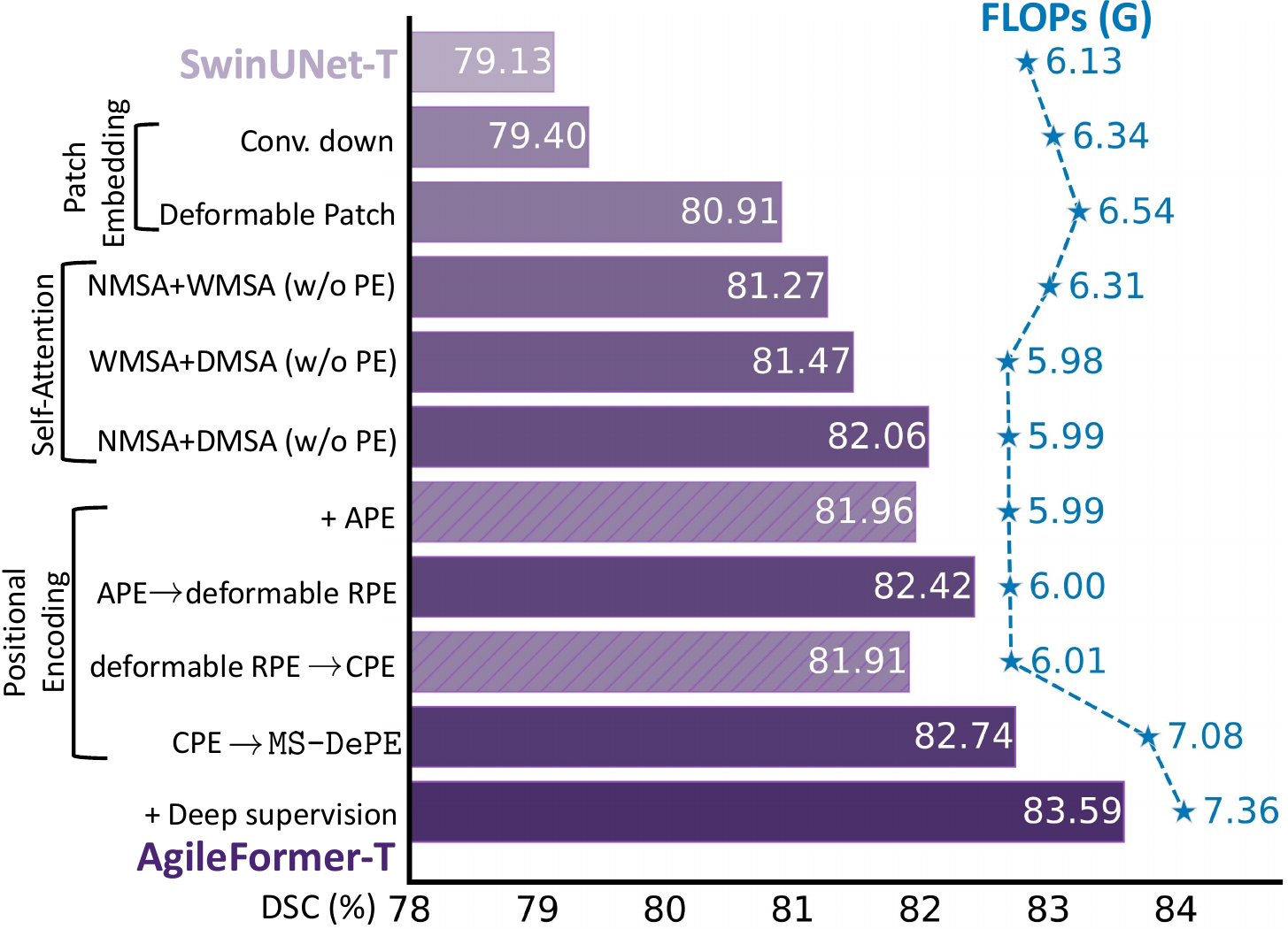}
    \caption{A roadmap going from a SwinUNet to the design of AgileFormer on Synapse dataset. From top
    to bottom, each row represents a model design variant, including patch embedding, self-attention, and positional encoding. The foreground bars represent
    DSC (\%) in the FLOP (G) regime of different design variants; a
hatched bar means the modification results in a performance drop.}
    \label{fig:roadmap}
\end{figure}

\begin{figure*}[!t]
    \centering
    \includegraphics[width=0.86\textwidth]{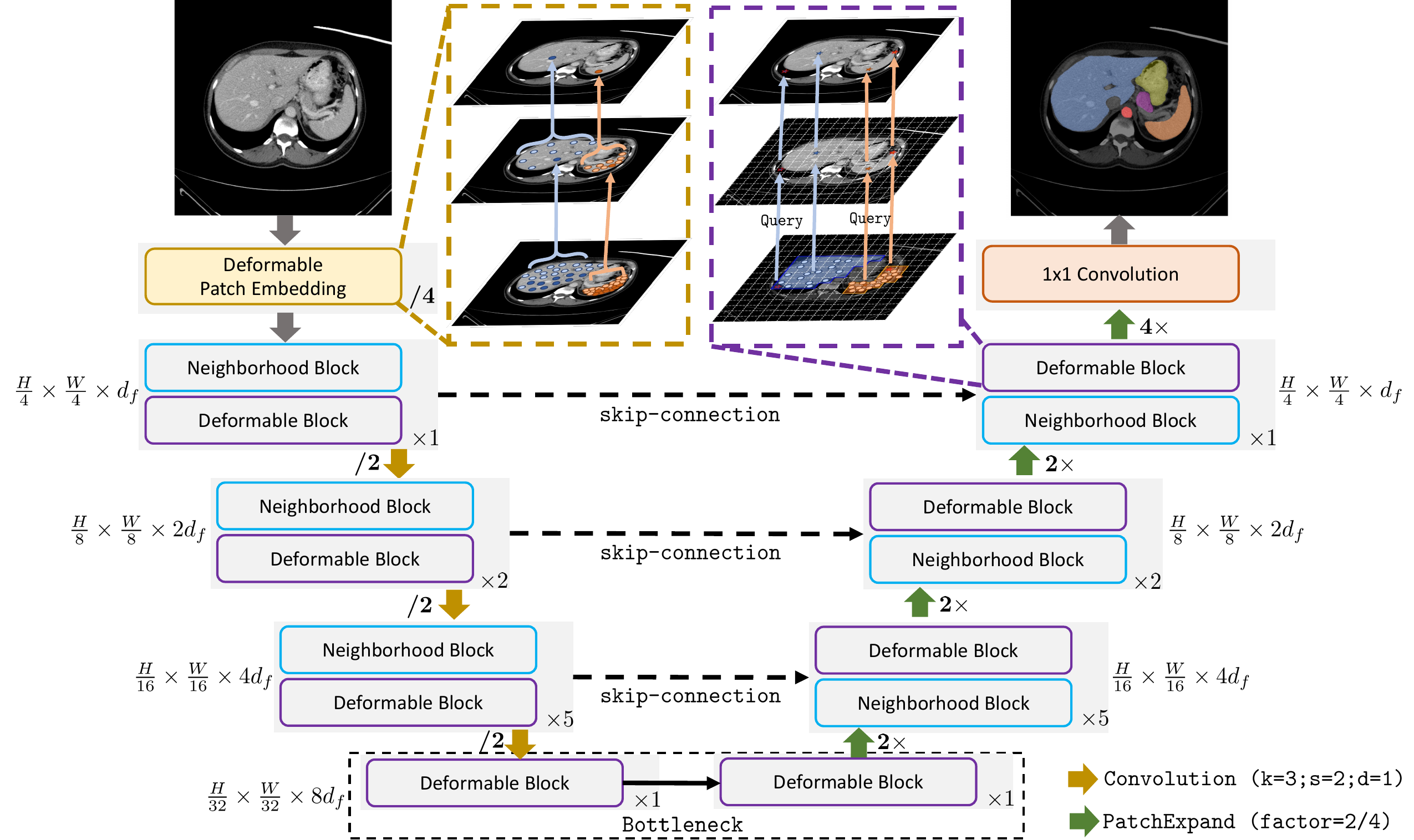}
    \caption{\textbf{The overview of the proposed AgileFormer}. AgileFormer is a U-shape vision transformer consisting of deformable patch embedding as well as neighborhood and deformable self-attention building block with a deformable positional encoding. For illustrative purposes, we take the 2D AgileFormer as an example. Please refer to ~\Cref{sec:3.4} for the detailed discussion on extending 2D AgileFormer to 3D AgileFormer for volumetric segmentation tasks. }
    \label{fig:network}
\end{figure*}

\section{Method}
In this section, we provide a roadmap going from the standard SwinUNet to the proposed AgileFormer (see~\Cref{fig:roadmap}). To make this paper self-contained, we first provide a brief introduction of the essential elements in ViT-UNets. A ViT-UNet (e.g., SwinUNet) is a U-shaped encoder-decoder network with skip connections, wherein the primary feature extractions in both the encoder and decoder are achieved by the self-attention mechanism. The building block of a standard ViT-UNet is composed of three fundamental components: patch embedding, self-attention, and positional encoding.
The patch embedding projects image patches into feature embeddings. More recent ViTs~\cite{xia2022vision,xia2023dat++} even treat downsampling (i.e., patch merging in a SwinUNet) as part of the patch embedding. In this paper, we will follow the same approach. The self-attention, which captures the dependencies between image patches, is used for the main feature extraction. For computational tractability and the need of locality, recent ViT-UNets use window-based self-attention mechanism~\cite{cao2022swin,heidari2023hiformer,huang2022swin,hatamizadeh2021swin}. Unlike convolution, self-attention discards the spatial correlation between image patches, which hinders the localization ability of pure ViT in segmentation tasks. The positional encoding is used to address this limitation.

\subsection{Deformable Patch Embedding}\label{sec:dpe}
The ViT-UNet starts by converting image patches into tokens. This process typically involves
partitioning an image into a sequence of non-overlapping $n \times n \ (\times n)$ square patches (e.g., $4 \times 4$ in SwinUNet~\cite{cao2022swin} for 2D segmentation tasks and $4 \times 4 \times 4$ in nnFormer for 3D segmentation tasks
). Subsequently, each of these patches is projected into a 1D feature vector. The main reason for performing a rigid (square) patch embedding is due to its simplicity, as it can be easily implemented as a standard convolution (kernel size $k=n$; stride $s=n$; dilation $d=1$). However, we argue that this rigid patch embedding is not optimal for a segmentation task for two main reasons. First, segmentation requires more precise pixel-level localization. The rigid patch embedding can only provide patch-level localization. 
Second, the shape and size of target objects in most medical image segmentation tasks vary significantly. 

\noindent \textbf{Deformable patch embedding.} To address the limitation of rigid patch embedding, we propose a deformable patch embedding (see~\Cref{fig:network}) by leveraging the deformable convolution~\cite{dai2017deformable}. The deformable convolution is defined as
\begin{equation}\label{eq:1}
    (\boldsymbol{f} \ast \boldsymbol{k})[p] = \sum_{p_k \in \Omega} \phi(\boldsymbol{f}; p + \Delta p + p_k ) \cdot \boldsymbol{k}[p_k],
\end{equation}
where $\boldsymbol{f} \in \mathbb{R}^{L\times d_f}$ is a $d_f$-dimensional feature map with a uniform grid of $L$ locations $p \in \mathbb{R}^{L \times D}$ ($D = 2$ for 2D; $D = 3$ for 3D). $\boldsymbol{k}$ is the convolutional kernel that operates on the grid $\Omega=[p_k]$ and defines the $k$-nearest (i.e., kernel size) neighboring locations of $p$. $[\Delta p]$ are offsets from which the irregular grid is sampled. The offsets are learned through a single convolutional layer $\Delta p = \operatorname{Conv}_{\text{offset}}(\boldsymbol{f})$. $\phi$ is a sampling function that performs bilinear/trilinear interpolation to sample locations $[p + \Delta p + p_k]$ in $\boldsymbol{f}$, as the offsets $\Delta p$ are typically fractional. 

\noindent \textbf{First patch embedding.} We replaced the single-layer rigid convolution patch embedding in a standard SwinUNet with two consecutive deformable convolutional layers ($k=3$; $s=n/2$; $d=1$). 
The rationale behind this is that the two consecutive overlapping deformable patch embeddings can extract better local representations, which compensates for the lack of locality in the self-attention. 

\noindent \textbf{Downsampling patch embedding.} We also replaced the patch merging in standard SwinUNet with convolutional downsampling via a single layer ($k=3$; $s=2$; $d=1$). We used overlapping kernels to better preserve localized patterns~\cite{huang2023vision,xia2022vision,zhou2021nnformer}, which also aligns with the overlapping deformable patch embedding.

\subsection{Spatially Dynamic Self-Attention}\label{sec:dmha}
The self-attention is the building block of a ViT UNet. Unlike convolution, self-attention does not enforce any spatial inductive bias but makes decisions purely by relying on dependencies (mainly similarities) between tokens while lacking the ability to capture spatially adaptive features for multi-class segmentation. Accordingly, we propose using a spatially dynamic self-attention module as the building block of ViT-UNet. This module is inspired from~\cite{xia2022vision}, and it includes deformable multi-head attention (DMSA)~\cite{xia2022vision} and neighborhood multi-head attention (NMSA)~\cite{hassani2023neighborhood}. 
The transformer block was constructed by alternating these two attention mechanisms (see~\Cref{fig:network}). We also distributed more computation to the third encoder block with a stage ratio of $[1:2:5:1]$, instead of $[2: 2: 2: 2]$ as used in~\cite{cao2022swin,zhou2021nnformer}. This is because the third layer of the encoder typically captures better feature representations than other layers~\cite{wang2022uctransnet}.

\subsubsection{Deformable Multi-head Self-Attention}
The deformable multi-head (i.e., $H$ head) attention~\cite{xia2022vision} for the $h$-th head is formulated as
\begin{equation}\label{eq:2}
    \operatorname{DMSA}_h(\boldsymbol{f}) = \operatorname{softmax}(\boldsymbol{Q}_h\Tilde{\boldsymbol{K}}_h^{\top} / \sqrt{d_k}) \Tilde{\boldsymbol{V}}_h
\end{equation}
where:
\begin{equation}\label{eq:3}
\begin{split}
     \boldsymbol{Q_h} = \boldsymbol{f} \boldsymbol{W}_h^{Q}, \quad \Tilde{\boldsymbol{K}}_h &= \Tilde{\boldsymbol{f}} \boldsymbol{W}_h^{K}, \quad  \Tilde{\boldsymbol{V}}_h = \Tilde{\boldsymbol{f}} \boldsymbol{W}_h^{V} \\ \text{with} \quad & \Tilde{\boldsymbol{f}} = \phi(\boldsymbol{f}; p + \Delta p_h).
\end{split}
\end{equation}
Here, we reuse the notation in \Cref{eq:1} with $p$ being a uniform grid of points, $\Delta p_h$ being the generated offsets for the $h$-th head, and $\phi$ being an interpolation function. 
$\{\boldsymbol{W}_h^Q, \boldsymbol{W}_h^K\} \in \mathbb{R}^{d_f \times d_k}$ and $\boldsymbol{W}_h^V \in \mathbb{R}^{d_f \times d_v}$ are trainable parameters, and $d_k$, $d_v$ are the hidden dimensions of linear projection of the key $\Tilde{\boldsymbol{K}}_h$ and the value $\Tilde{\boldsymbol{V}}_h$ in DMSA.
The offsets in DMSA are also generated by passing the query through a convolutional layer $\Delta p = \operatorname{Conv}_{\text{offset}}(\boldsymbol{Q}_h)$. The resulting irregularly sampled feature map is denoted as $\Tilde{\boldsymbol{f}}$ in \Cref{eq:3}.
Similar to deformable convolution, the irregularly sampled feature map is then applied to self-attention by using irregularly sampled key $\Tilde{\boldsymbol{K}}_h$ and value $\Tilde{\boldsymbol{V}}_h$ (see~\Cref{eq:2}). 

\subsubsection{Neighborhood Multi-head Self-Attention} In contrast to standard self-attention, which computes the similarity of each element at a given position $p$ on a feature map $\boldsymbol{f}$ with every other element, the construction of neighborhood attention~\cite{hassani2023neighborhood} only leverages the information from the $k$-nearest neighbors around location $p$. We reuse the notation $k$ in \Cref{eq:2}, as the NMSA operates like a convolution. As a result, neighborhood attention reduces the quadratic computational complexity in standard self-attention to approximately linear w.r.t. the spatial dimension of $\boldsymbol{f}$, as $k$ is typically small (e.g., $k = 7 \times 7 \ (\times 7)$). Furthermore, this reintroduces local operations into self-attention, allowing for translational equivariance, and thereby enhancing the ability to better preserve local information.   
Following the notation in \Cref{eq:2}, the neighborhood multi-head  attention at position $p_l$ is computed as
\begin{equation} \nonumber
\begin{split}
\operatorname{NMSA}_h(&\boldsymbol{f}[p_l]) = \operatorname{softmax}(\boldsymbol{Q}_h[p_l] \hat{\boldsymbol{K}}_h[p_l]^{\top} / \sqrt{d_k})  \hat{\boldsymbol{V}}_h[p_l] \\
\end{split}
\end{equation}
where:
\begin{equation} \nonumber
\begin{split}
      \hat{\boldsymbol{K}}_h[p_l] &= [\boldsymbol{K}_h[p_{l1}], \cdots, \boldsymbol{K}_h[p_{lk}]], \quad \boldsymbol{K}_h = \boldsymbol{f}\boldsymbol{W}_h^{K}\\
     \hat{\boldsymbol{V}}_h[p_l] &= [\boldsymbol{V}_h[p_{l1}]^{\top}, \cdots, \boldsymbol{V}_h[p_{lk}]^{\top}]^{\top}, \quad \boldsymbol{V}_h = \boldsymbol{f}\boldsymbol{W}_h^{V}.
\end{split}
\end{equation}
Here, $[p_{lk}]$ denotes the $k$-th neighboring locations for a given location $p_l$. It is worth noting that the dimension of the resulting attention is $\mathbb{R}^{L \times K}$ with $K = k \times k~(\times k)$, instead of $\mathbb{R}^{L \times L}$ in standard self-attention and window-attention.

\subsection{Multi-scale Deformable Positional Encoding}\label{sec:dposenc}
\begin{figure}[!t]
    \centering
    \includegraphics[width=1\columnwidth]{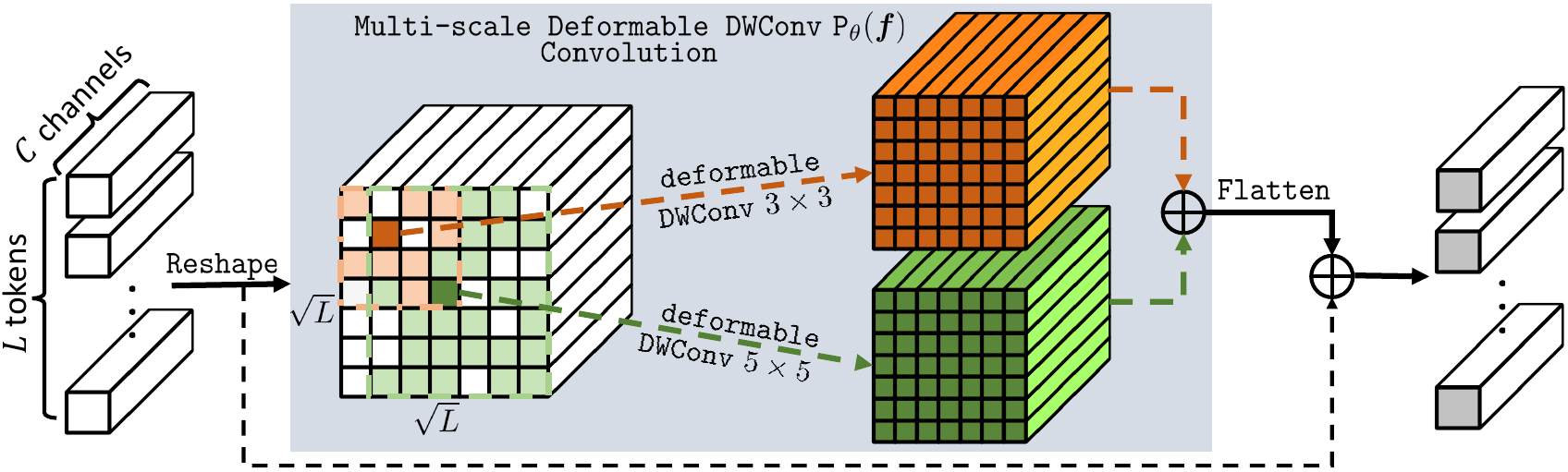}
    \caption{The proposed multi-scale deformable positional encoding ($\operatorname{MS-DePE}$) for irregularly sampled grids in deformable multi-head self-attention. For illustrative purposes, we take the 2D model as an example, where 2D deformable depth-wise convolution is used. In the 3D model, the 2D deformable depth-wise convolution should be replaced with its 3D counterpart.}
    \label{fig:DePE}
\end{figure}

The design of positional encoding (PE) is barely explored by previous ViT UNets. Most ViT-UNets either disregard positional encoding~\cite{chen2021transunet} or inherit PEs from their ancestor models~\cite{wang2021transbts,cao2022swin,heidari2023hiformer}. 
Specifically, \cite{wang2021transbts} used an absolute PE (APE), which assigns an absolute value for each token. While the others~\cite{cao2022swin,heidari2023hiformer,zhou2021nnformer} used a relative PE (RPE)~\cite{shaw2018self} to encode the relative positions between tokens.
However, these are designed for 1D signals while neglecting spatial correlations. Consequently, they are not well adapted for modeling 2D/3D signals with spatial correlations. Recently, conditional PE (CPE)~\cite{chu2022conditional} was designed for vision tasks offering the ability of both APE and RPE at the same time. 
More importantly, standard APE, RPE, and CPE are not directly applicable to irregularly sampled grids, as all of them are formulated in a rigid grid. In accordance with the irregularly sampled DMSA presented in this paper, we propose a multi-scale deformable positional encoding ($\operatorname{MS-DePE}$) designed to encode irregularly sampled positional grids across various scales (see~\Cref{fig:DePE}).

The proposed $\operatorname{MS-DePE}$ is formulated in a conditional positional encoding fashion~\cite{chu2022conditional} as follows: 
\begin{equation}\nonumber
    \operatorname{MS-DePE}(\boldsymbol{f}) = \boldsymbol{f} + \operatorname{P}_{\theta}(\boldsymbol{f}),
\end{equation}
where $\boldsymbol{f}$ is the input feature map, and $\texttt{P}_{\theta}$ denotes the trainable positional encoding layer parameterized by $\theta$. $\texttt{P}_{\theta}$ is implemented as multi-scale deformable depth-wise convolutional layers with different kernel sizes (i.e., $3\times 3 \ (\times 3)$ and $5\times 5 \ (\times 5)$). 
For this purpose, we first recover the spatial resolution of the feature map $\boldsymbol{f}$ before passing it through $\texttt{P}_{\theta}$. After applying $\texttt{P}_{\theta}$, we reshape the feature map back to its original shape (see Fig.~\ref{fig:DePE} for details).

\begin{table*}[!t]
\caption{Performance comparison with 2D and 3D segmentation networks in the multi-organ segmentation using the Synapse multi-organ segmentation dataset. The best result for 2D and 3D models within each column is highlighted by \textbf{bold}, and the second-best is highlighted with an \underline{underline}. $^\dagger$: models implemented by us. The other benchmarks were taken from~\cite{chen2021transunet,cao2022swin,rahman2024multi,wang2022mixed,heidari2023hiformer,huang2022missformer,yan2022after,rahman2023medical,zhou2021nnformer}. }
\centering
\begin{threeparttable}
\resizebox{0.95\textwidth}{!}{
\begin{tabular}{cl|cc|cccccccc}
\toprule
& Methods & Avg. DSC$\uparrow$ & HD95 $\downarrow$  & Aorta & Gallbladder & Kidney(L) & Kidney(R) & Liver & Pancreas & Spleen & Stomach \\
\midrule
&U-Net~\cite{ronneberger2015u}  & 76.85 & 39.70 & 89.07 & 69.72 & 77.77 & 68.60 & 93.43 & 53.98 & 86.67 & 75.58 \\
&Att-UNet~\cite{oktay2018attention}  & 77.77 & 36.02 & \textbf{89.55} & 68.88 & 77.98 & 71.11 & 93.57 & 58.04 & 87.30 & 75.75 \\
&TransUNet~\cite{chen2021transunet} & 77.48 &31.69  & 87.23 & 63.13 & 81.87 & 77.02 & 94.08 & 55.86 & 85.08 & 75.62 \\ 
&MixedUNet~\cite{wang2022mixed}  & 78.59 & 26.59 & 87.92 & 64.99 & 81.47 & 77.29 & 93.06 & 59.46 & 87.75 & 76.81 \\ 
&$^\dagger$CoTr~\cite{xie2021cotr}  & 78.56 &24.05 & 87.09 & 65.37 & 86.19 & 80.32 & 94.22 & 52.28 & 87.01 & 76.00 \\
&Hiformer~\cite{heidari2023hiformer}  &80.69 &19.14 & 87.03& 68.61 & 84.23 & 78.37 & 94.07 & 60.77 & 90.44 & 82.03\\
& PVT-GCASCADE~\cite{rahman2024g} & 83.28 & 15.83 & 88.05 & 74.81 & 88.02 & 84.83 & 95.38 & 69.73 & 91.92 & 83.63 \\
&$^{\nabla}$Trans-CASCADE~\cite{rahman2023medical}  & 82.68 & 17.34 & 86.63 & 68.48 & 87.66 & 84.56 & 94.43 & 65.33 & 90.79 & 83.69\\
\multirow{3}{*}[0.2cm]{\rotatebox[origin=c]{90}{\makecell{\textbf{2D Models}}}}&R50 ViT~\cite{chen2021transunet}  & 71.29 & 32.87 & 73.73 & 55.13 & 75.80 & 72.20 & 91.51 & 45.99 & 81.99 & 73.95 \\
&SwinUNet~\cite{cao2022swin}  & 79.13 & 21.55 & 85.47 & 66.53 & 83.28 & 79.61 & 94.29 & 56.58 & 90.66 & 76.60 \\
&TransDeepLab~\cite{azad2022transdeeplab}  & 80.16 &21.25 & 86.04 & 69.16 & 84.08& 79.88 & 93.53 & 61.19 & 89.00 & 78.40 \\
& $^\dagger$SDAUT~\cite{huang2022swin} & 80.67 & 25.59 & 87.03  & 69.30  & 81.87 & 80.20 & 94.91 & 64.56 & 89.54 & 77.91  \\
& CATFormer~\cite{you2022class} & 82.17 &16.20 & 88.98 & 67.16 & 85.72 & 81.69 & 95.34 & 66.53 & 90.74 & 81.20 \\
& MissFormer~\cite{huang2022missformer} & 81.96 & 18.20 & 86.99 & 68.65 & 85.21 & 82.00 & 94.41 & 65.67 & 91.92 & 80.81 \\
& $^{\nabla}$MERIT~\cite{rahman2024multi} & \underline{84.90} & \textbf{13.22} & 87.71 & 74.40 & \underline{87.79} & \underline{84.85} & 95.26 & \textbf{71.81} & \underline{92.01} & 85.38 \\ 
\rowcolor{blue!8}
& \textbf{AgileFormer-T w/o DS}  & 82.74 & 19.28 & 88.08 & \underline{75.16} & 82.41 & 81.36 & 95.09 & 67.23  & 90.94 & 81.61 \\
\rowcolor{blue!8}
& \textbf{AgileFormer-T w/ DS}   & 83.59 & \underline{15.09} &  88.81 & 74.43 & 84.61 & 82.78 & \underline{95.48} & 69.45  & 90.14 & 83.05 \\
\rowcolor{blue!8}
& \textbf{AgileFormer-B w/o DS}   & 84.14 & 15.27 & 87.76 & 74.71 & 86.69 & 83.81 & 95.31 & 68.28 & 91.00 & \underline{85.58} \\
\rowcolor{blue!8}
& \textbf{AgileFormer-B w/ DS}  & \textbf{85.74}$^*$ & 18.70  & \underline{89.11} & \textbf{77.89} & \textbf{88.83} & \textbf{85.00} & \textbf{95.64} & \underline{71.62} & \textbf{92.20} & \textbf{85.63} \\
\midrule
\multirow{3}{*}[0.0cm]{\rotatebox[origin=c]{90}{\makecell{\textbf{3D Models}}}} 
& nnUNet~\cite{isensee2021nnu} & \underline{86.99} & 10.78 & \textbf{93.01} & \underline{71.77} & 85.57 & \textbf{88.18} & \textbf{97.23} & \underline{83.01} & 91.86 & \underline{85.26} \\
& UNETR~\cite{hatamizadeh2022unetr} & 79.56 & 22.97 & 89.99 & 60.56 & 85.66 & 84.80 & 94.46 & 59.25 & 87.81 & 73.99  \\
& SwinUNETR~\cite{hatamizadeh2021swin} & 83.48 & \underline{10.55} & 91.12 & 66.54 & \underline{86.99} &  86.26 &  95.72 &  68.80 & \underline{95.37} & 77.01  \\
& nnFormer~\cite{zhou2021nnformer} & 86.57 & 10.63 & \underline{92.04} & 70.17 & 86.57 & 86.25 & 96.84 & \textbf{83.35} & 90.51 & \textbf{86.83} \\
\rowcolor{blue!8}
& \textbf{AgileFormer-T w/ DS} & \textbf{87.43}$^*$ & \textbf{7.81}$^*$ & 91.80 & \textbf{73.02} & \textbf{87.63} & \underline{87.49} & \underline{97.06} & 82.05 & \textbf{95.88} & 84.48 \\
\bottomrule
\end{tabular}}
\begin{tablenotes}
\item \scriptsize $^*$: $p<0.05$; with Wilcoxon signed rank test to the second-best model.  $^{\nabla}$: Large models that have similar parameters to the AgileFormer-B model (\Cref{fig:graphic_abstract} (\textbf{Right})), where \\ \indent deep supervision is also applied. \textbf{Note}: All 2D AgileFormer results reported in this paper use the checkpoint from the last training epoch. If using the testing set to select the \\ \indent best model (e.g., as done in the code of MERIT, Trans-CASCADE, and PVT-GCASCADE), 2D Agileformer-B can achieve a DSC (\%) and HD95 of 86.11 and 12.88.
\end{tablenotes}
\end{threeparttable}
\label{tab:synapse}
\end{table*}

\subsection{Model Configurations}\label{sec:3.4}
\noindent \textbf{2D AgileFormer.} 
Similar to~\cite{chen2021transunet,cao2022swin,rahman2024multi}, we developed two variants of AgileFormer by varying the embedding dimension ($d_f$) and the number of heads ($H$), while keeping the main structures unchanged (i.e., number of transformer blocks in the encoder/decoder): AgileFormer-T(iny) ($d_f=[64, 128, 256, 512]$; $H=[2, 4, 8, 16]$) and AgileFormer-B(ase) ($d_f=[128, 256, 512, 1024]$; $H=[4, 8, 16, 32]$). We also incorporated deep supervision (DS) outlined in~\cite{isensee2018brain,zhou2021nnformer} into the proposed method.

\noindent \textbf{3D AgileFormer.} For 3D models, we only evaluated the tiny model (i.e., AgileFormer-T) that has a comparable size to other baseline models. We also used the same $d_f$ and $H$ as in the 2D model. However, we made a few modifications to accommodate the 3D input. First, we replaced 2D convolutions with 3D convolutions to generate the offset points (Conv$_{\text{offset}}(\boldsymbol{Q}_h)$). Second, we sampled positions in the 3D grid based on the generated offset points. 
Third, for the \texttt{key} and \texttt{query} projection in DMSA, we kept 2D $1 \times 1$ convolutional projection. This is because 2D and 3D convolutions with a kernel size of 1 essentially perform the weighted summation over channels. Accordingly, we first reshaped the 5D tensor of size \texttt{$[B \times C \times D \times H \times W]$} into a 4D tensor \texttt{$[B \times C \times 1 \times (D*H*W)]$} and then performed a 2D $1\times 1$ convolution. After finishing the computation of self-attention, we reshaped the 4D tensor back to 5D to restore its spatial dimension. Fourth, we replaced all 2D convolutions in the deformable patching embedding and multi-scale deformable positional encoding with 3D convolutions. 

\section{Experiments and Results}
\subsection{Experimental design}
\noindent \textbf{Dataset.} In line with previous works \cite{chen2021transunet,cao2022swin,zhou2021nnformer}, we validated the proposed method on three publicly available medical image segmentation datasets: the Synapse multi-organ dataset~\cite{landman2015miccai}, the Automated Cardiac Diagnosis Challenge (ACDC) dataset~\cite{bernard2018deep}, and the brain tumor segmentation dataset from the Decathlon challenge~\cite{antonelli2022medical}. The Synapse multi-organ dataset includes 30 3D abdominal CT scans with corresponding segmentation masks of 13 organs. For brevity, we refer to this dataset as Synapse hereafter. As consistent with prior works, we report the performance on 8 abdominal organs (i.e., aorta, gallbladder, kidney left, kidney right, liver, pancreas, spleen, stomach). The ACDC dataset consists of 100 3D cardiac MRI scans, each with a segmentation mask that features the right ventricle (RV), myocardium (Myo), and left ventricle (LV). The Decathlon brain tumor dataset comprises 484 3D multi-modal brain tumor MRI scans with segmentation masks delineating enhancing tumor, non-enhancing tumor, and edema. To be consistent with previous works~\cite{hatamizadeh2022unetr,zhou2021nnformer}, we reported the results of the whole tumor (WT), enhancing tumor (ET) and tumor core (TC). 

\begin{table}[!t]
    \centering
    \caption{Performance comparison with 2D and 3D segmentation models in cardiac MRI segmentation using the ACDC dataset. $^\dagger$: models implemented by us. The other benchmarks were taken from~\cite{chen2021transunet,cao2022swin,wang2022mixed,huang2022missformer,rahman2023medical,rahman2024multi,zhou2021nnformer}.}
    \begin{threeparttable}
    \resizebox{1\columnwidth}{!}{
    \begin{tabular}{cl|c|ccc}
    \toprule
        & Methods & Avg. DSC$\uparrow$ & RV & Myo & LV \\
    \midrule
        &TransUNet~\cite{chen2021transunet} & 89.71 & 88.86 & 84.53 & 95.73 \\
        & $^\dagger$CoTr~\cite{xie2021cotr} & 90.52 & 87.81 & 88.44 & 95.29 \\   
        & SwinUNet~\cite{cao2022swin} & 90.00 & 88.55 & 85.62 & 95.83 \\
        & MixedUNet~\cite{wang2022mixed} & 90.43 & 86.64 & 89.04 & 95.62 \\
        & MissFormer~\cite{huang2022missformer} & 90.86 & 89.55 & 88.04 & 94.99 \\
        \multirow{3}{*}[0.8cm]{\rotatebox[origin=c]{90}{\makecell{\textbf{2D Models}}}} & $^\dagger$SDAUT~\cite{huang2022swin} & 91.08 & 89.37 & 88.58 & 95.28\\
        & $^\nabla$PVT-CASCADDE~\cite{rahman2023medical} & 91.46 & 88.00 & 89.97 & 95.50\\
        & $^\nabla$Trans-CASCADE~\cite{rahman2023medical} & 91.63 & 89.14 & 90.25 & 95.50 \\
        & $^\dagger$MERIT~\cite{rahman2024multi} & \underline{91.81} & \underline{90.44} & \underline{89.12} & \underline{95.85} \\
         \rowcolor{blue!8}
        & \textbf{AgileFormer-T w/ DS} &  91.76 & 89.80 & 89.71 & 95.77\\
         \rowcolor{blue!8}
        & \textbf{AgileFormer-B w/ DS} & \textbf{92.55}$^*$ & \textbf{91.05} & \textbf{90.40} & \textbf{96.19}\\
         \midrule
         \multirow{3}{*}[0.0cm]{\rotatebox[origin=c]{90}{\makecell{\textbf{3D Models}}}} & UNETR~\cite{hatamizadeh2022unetr} & 88.61 & 85.29 & 86.52 & 94.02 \\
         & UNETR++~\cite{shaker2024unetr++} & 92.83 & 91.89 & 90.61 & 96.00 \\
         & PHTrans & 91.79 & 90.13 & \underline{89.48} & \underline{95.76} \\ 
         & nnUNet~\cite{isensee2021nnu} & 91.61 & 90.24 & 89.28 & 95.36 \\
         & $^\dagger$nnFormer~\cite{zhou2021nnformer} & \underline{91.84} & \textbf{90.91} & 89.44 & 95.17 \\
         \rowcolor{blue!8}
        & \textbf{AgileFormer-T w/ DS} & \textbf{92.07}$^*$ & \underline{90.59} & \textbf{89.80} & \textbf{95.81} \\
         \bottomrule
    \end{tabular}}
    \begin{tablenotes}
    \item \scriptsize $^*$: $p<0.05$; with the paired t-test to the second-best model. We reproduce the \\ \indent results of nnFormer by taking the model weights released by the authors in~\cite{zhou2021nnformer}. 
    \end{tablenotes}
    \end{threeparttable}
    \label{tab:ACDC}
\end{table}
\begin{table}[!t]
    \centering
        \caption{Performance comparison with 3D segmentation models in brain tumor segmentation using the Decathlon dataset. $^\dagger$: models implemented by us. 
        }
        \begin{threeparttable}
        \resizebox{1\columnwidth}{!}{
           \begin{tabular}{l|c|ccc}
    \toprule
        Methods & Avg. DSC$\uparrow$  & WT & ET & TC \\
    \midrule    
    $^\dagger$nnUNet~\cite{isensee2021nnu} & 85.2 & 91.0 & 80.3 & 84.1 \\
    $^\dagger$MedNext~\cite{roy2023mednext} & \underline{85.4} & 90.3 & \textbf{81.5} & \underline{84.5}  \\
    \midrule
    $^\dagger$TransUNet~\cite{chen2021transunet} & 84.6 & 90.5 & 79.9 & 83.4 \\
         $^\dagger$TransBTS~\cite{wang2021transbts} & 84.2 & 90.2 & 79.6 & 82.9 \\
         $^\dagger$CoTr~\cite{xie2021cotr} & 84.1 & \underline{91.2} & 78.7 & 83.2 \\
         $^\dagger$UNETR~\cite{hatamizadeh2022unetr} & 84.1 & 90.5 & 79.3 & 82.4 \\
         $^\dagger$UNETR++~\cite{shaker2024unetr++} & 85.2 & 91.0  & 80.1  & 84.2 \\
         $^\dagger$SwinUNETR~\cite{hatamizadeh2021swin} & 84.9  & 90.7 & 80.5 & 83.5 \\$^\dagger$nnFormer~\cite{zhou2021nnformer} & 84.9 & \textbf{91.4} & 79.2 & 84.0 \\
         \midrule
         \rowcolor{blue!8}
          \textbf{AgileFormer-T w/ DS} & \textbf{85.7$^*$} & \underline{91.2} & \underline{80.8} & \textbf{85.1} \\
         \bottomrule
    \end{tabular}}
    \begin{tablenotes}
    \item \scriptsize $^*$: $p<0.05$; with the paired t-test to the second-best model. 
    \end{tablenotes}
    \end{threeparttable}
    \label{tab:brain_tumor}
\end{table}

\noindent \textbf{Experimental setup and evaluation metric.}
We evaluated the performance of both 2D AgileFormer (trained using slices) and 3D AgileFormer (trained using volumes), as previous works involved both 2D and 3D models.
For 2D AgileFormer, we adhered to the experimental protocols outlined in~\cite{chen2021transunet,cao2022swin} for the Synapse and ACDC datasets, including training/testing partitioning, input image size (i.e., $224 \times 224$), data augmentations, model selection, and evaluations.
For the 3D AgileFormer, we followed the experimental protocols specified in~\cite{zhou2021nnformer} for a fair comparison with the previous literature. The input image sizes for Synapse, ACDC, and Decathlon brain tumor datasets were set to $64 \times 128 \times 128$, $14 \times 160 \times 160$, and $96 \times 96 \times 96$~\footnote{We used an input size of $96 \times 96 \times 96$ instead of $128 \times 128 \times 128$ in nnFormer~\cite{zhou2021nnformer} for brain tumor segmentation. This is because most baseline methods~\cite{chen2021transunet, hatamizadeh2022unetr,hatamizadeh2021swin,wang2021transbts,xie2021cotr} in nnFormer~\cite{zhou2021nnformer} were implemented in $96 \times 96 \times 96$, and hence show a huge performance difference.}, respectively.  
We evaluated the segmentation performance using the dice similarity coefficient (DSC) and 95\% Hausdorff
Distance (HD95). Statistical significance between the average DSC of the best and second-best model was estimated using the paired t-test. We emphasize that all competing methods applied on the Synapse, ACDC, and Decathlon datasets followed the same experimental protocols.

\noindent \textbf{Implementation details.} 
All 2D and 3D models were trained using a combination of DSC and cross-entropy loss~\cite{cao2022swin,zhou2021nnformer}, employing an AdamW~\cite{loshchilov2017decoupled} optimizer with cosine learning rate decay. We initialized model parameters with ImageNet pre-trained weights. Since different hyperparameter settings were used for 2D and 3D experiments, we kindly direct readers to \texttt{Appendix B} for the detailed model and hyperparameter configurations for each task.
All experiments were implemented using \texttt{PyTorch} and were performed on a Nvidia V100 GPU with 32GB memory. 

\subsection{Main results}

\begin{figure}[!t]
    \centering
    \includegraphics[width=1.0\columnwidth]{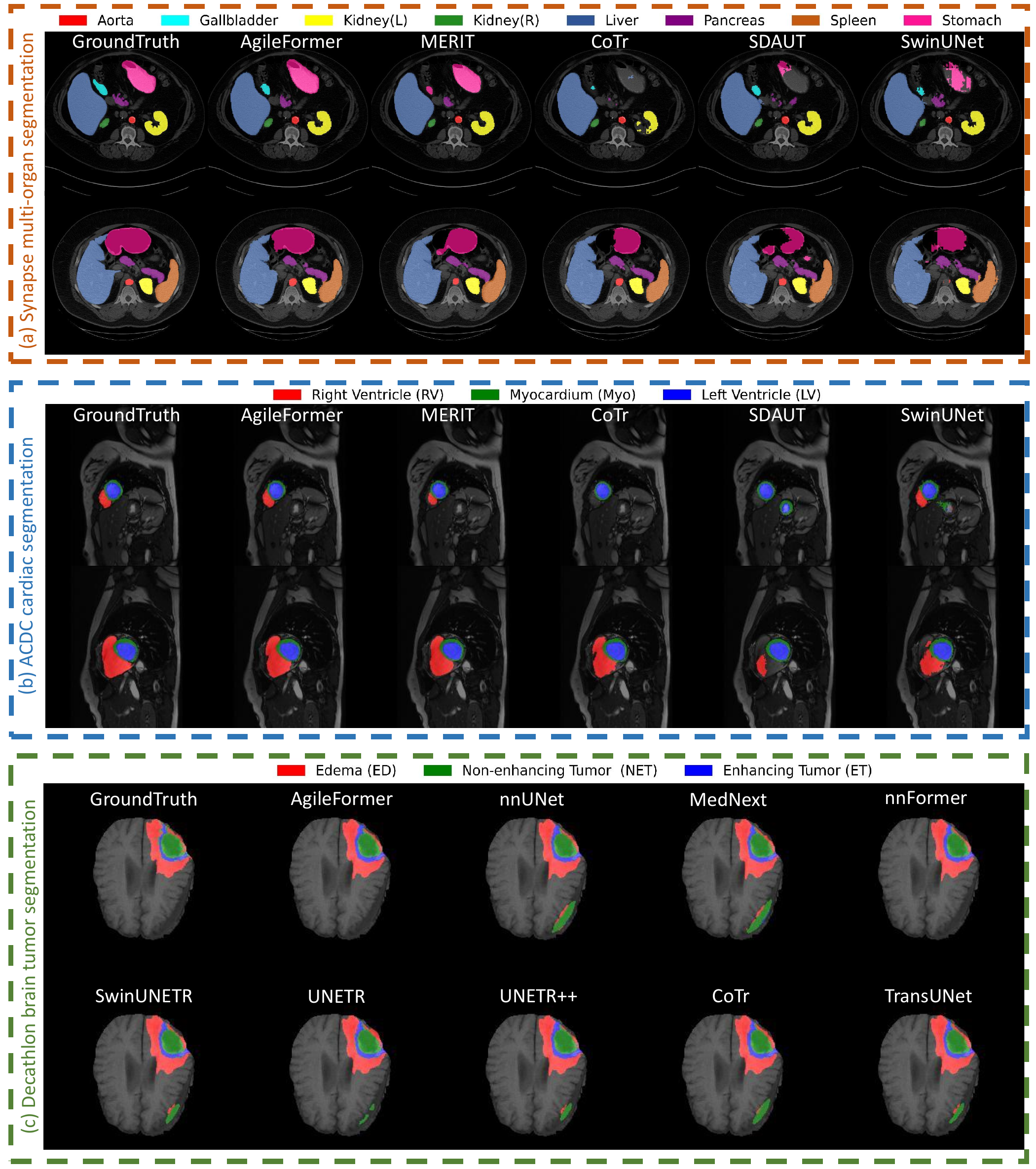}
    \caption{Visual comparison of (a) multi-organ, (b) cardiac, and (c) brain tumor segmentation across Synapse, ACDC, and Decathlon datasets. AgileFormer showed the best qualitative segmentation quality across all three segmentation tasks. }
    \label{fig:segm_vis}
\end{figure}

\noindent \textbf{Results of 2D Models.}
2D AgileFormer performed better than the second-best 2D models (i.e., MERIT~\cite{rahman2024multi} for both tasks) by 0.84 and 0.23, in average DSC (\%) on the Synapse and ACDC datasets, respectively. This improvement was shown to be statistically significant.
Specifically, the proposed method achieved an average DSC (\%) of 85.74, surpassing the other baseline methods in segmenting 6 out of 8 organs on the Synapse dataset (see~\Cref{tab:synapse}; \textbf{2D Models}). For the remaining two organs, the proposed method was ranked the second best. 
One the ACDC dataset, AgileFormer achieved an average DSC (\%) of 92.55, surpassing all the baseline methods in segmenting RV, Myo, and LV (see~\Cref{tab:ACDC}; \textbf{2D Models}). 
Qualitative results also supported the superiority of AgileFormer. While AgileFormer accurately segmented structures of interest, other methods either under- or over-segmented certain regions (see ~\Cref{fig:segm_vis} (a) and (b)).


\noindent \textbf{Results of 3D Models.}
When comparing 3D models, 3D AgileFormer also outperformed other state-of-the-art segmentation methods across all three segmentation tasks. Specifically, AgileFormer achieved an average DSC (\%) of 87.43 and an average HD95 (mm) of 7.81, surpassing nnFormer by 0.86 and 2.82 in DSC and HD95, respectively (\Cref{tab:synapse}; \textbf{3D Models}). On the ACDC dataset, AgileFormer achieved an average DSC (\%) of 92.07, outperforming nnFormer by 0.23 (\Cref{tab:ACDC}; \textbf{3D Models}).
On the brain tumor segmentation task, the proposed method achieved an average DSC (\%) of 85.7, surpassing nnFormer by 0.8 (\Cref{tab:brain_tumor}). In addition, AgileFormer showed high qualitative segmentation quality on 3D segmentation tasks (see  \Cref{fig:graphic_abstract} and~\Cref{fig:segm_vis} (c)). In summary, the proposed AgileFormer was more effective on multi-organ and brain tumor segmentation tasks, where target objects exhibit more heterogeneous appearances than those in cardiac segmentation.


Additional qualitative results on Synapse produced by 3D models are shown in~\Cref{fig:vis}, where AgileFormer showcased accurate segmentation compared to other state-of-the-art methods. Some failure cases on Synapse are shown in~\Cref{fig:failure_synapse}, where even methods that can extract spatially dynamic features (including AgileFormer) cannot accurately segment highly irregular organs such as Gallbladder and Pancreas. Similarly,~\Cref{fig:failure_tumor} shows failure cases on brain tumor segmentation tasks, where tissue classes with highly irregular appearance cannot be accurately segmented by most methods, including AgileFormer.

\begin{figure}[!t]
    \centering
    \includegraphics[width=1\columnwidth]{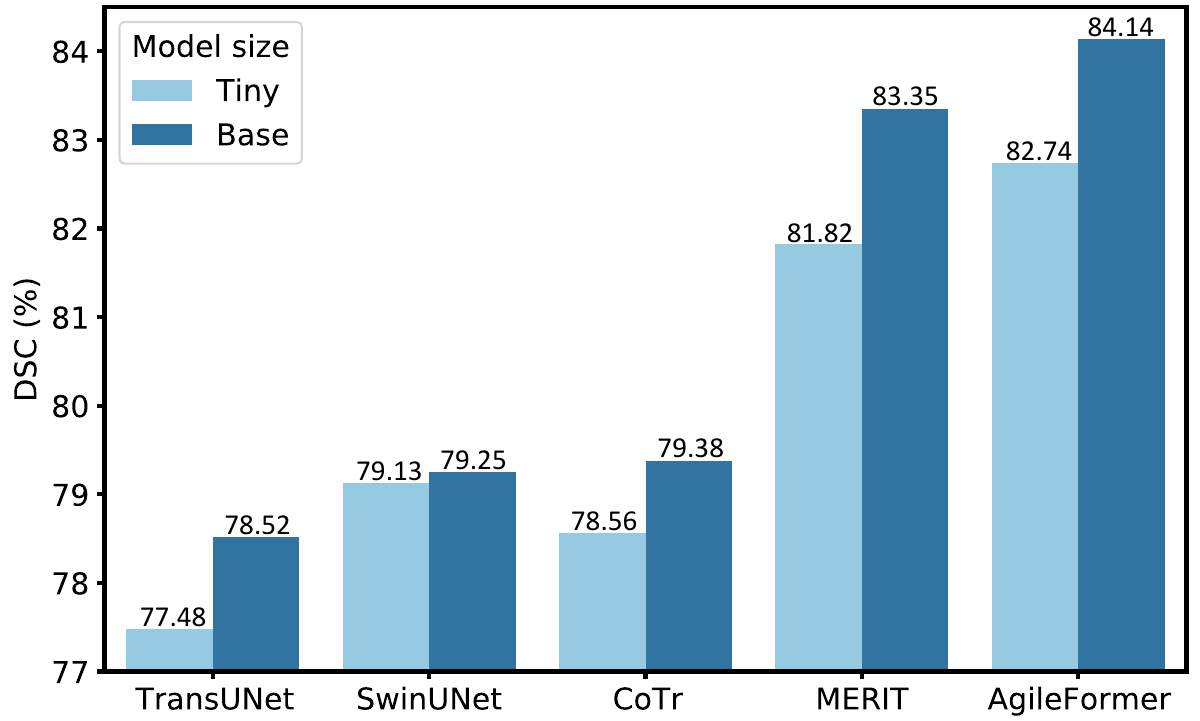}
    \caption{Comparison of model scalability on the Synapse dataset. The base model is almost four times larger than the tiny model. The proposed AgileFormer demonstrated exceptional scalability when adding parameters compared to other methods.}
    \label{fig:scalability}
\end{figure}

\subsection{Ablations on model design variants}
We conducted ablation studies on the model design variants on the Synapse dataset using 2D AgileFormer, including patch embedding and different choices of spatially dynamic attention and positional encoding (see~\Cref{fig:roadmap}). First, replacing patch merging with convolutional downsampling led to a small performance gain of $0.34\%$ in average DSC.
Then, adding deformable embedding led to a performance gain of $1.9\%$ in DSC. Second, we ablated different choices of spatially dynamic self-attention by alternating window/deformable/neighborhood attention. We removed positional encoding to eliminate its effect for now. Alternating window attention (WMSA) with both DMSA and NSMA led to a performance gain of $0.7\%$ (DMSA) and $0.4\%$ (NSMA), respectively. Alternating NMSA and DMSA resulted in a performance gain of $1.4\%$. Third, we brought back positional encoding, demonstrating that not all PEs can lead to a performance gain, e.g., adding APE and CPE even led to a performance drop. This is because these two PEs are not designed for irregularly sampled grids introduced by DMSA. Instead, the proposed $\operatorname{MS-DePE}$ improved performance by $0.8\%$. Notably, the proposed AgileFormer did not lead to a huge computational burden, with only a 1.1\% increase in the number of parameters and a 15\% rise in floating point operations, compared to SwinUNet.

\subsection{Model scalability}
We compared the scaling behavior of the proposed method with other ViT-UNets on the Synapse dataset using 2D AgileFormer. 
As shown in~\Cref{fig:scalability}, the proposed method demonstrated improved performance with growing model sizes for the Synapse multi-organ segmentation task, where different organs exhibited a lot of variability in shape and size. Specifically, the performance was improved by 2.15 in DSC (\%) going from AgileFormer-T to AgileFormer-B, with parameters increasing approximately fourfold from 28.85M to 123.47M. 
We also observed that
the methods (i.e., CoTr, MERIT, and AgileFormer) can capture spatially varying representations generally scaled better than standard ViT-UNets (i.e., TransUNet and SwinUNet). However, CoTr did not scale as well as MERIT and AgileFormer. We hypothesized this might be attributed to the fact that the main backbone of CoTr was still a CNN. The deformable module in CoTr was only placed in the bottleneck. 

\subsection{Discussion}
\noindent\textbf{Failure cases}: Although AgileFormer demonstrated improvements over previous state-fo-the-art segmentation methods quantitatively and qualitatively, it still has limitations on highly irregular and unstructured objects (e.g., on Pancreas and Gallbladder (see~\Cref{fig:failure_synapse}) and some tissue classes in brain tumor segmentation (see~\Cref{fig:failure_tumor})). This suggests that there is still room for improvement in future work with more powerful spatially varying components; whereas our investigation revealed the importance of incorporating spatially varying components into UNet designs.
\\
\noindent\textbf{Potential trade-offs between different organs}: First, we observed that AgileFormer showed more promising results on challenging small and irregular organs (e.g., Gallbladder Aorta, and Pancreas) but also achieved the best performance on most large organs (see Table 1). Second, we observed that many ViT-UNets can perform equally well on the larger organs (e.g., Liver and Spleen). However, the pure CNN-based methods did not work well on the larger organs. We conjectured that this phenomena was attributed to the fact that the self-attention mechanism in ViT-UNets better capture global context than convolution in CNNs (which is typically limited by the receptive field).
Third, CNN-based models (U-Net~\cite{ronneberger2015u} and Att-UNet~\cite{oktay2018attention}) biased toward small organs (e.g., Aorta) due to the inherent locality of convolution in capture small objects. In contrast, the rigid patching embedding in ViT-UNets may fail to capture such localized structures and hence showed inferior performance on small organs. Notably, our investigation revealed that replacing the rigid patch embedding with a deformable patch embedding dramatically improved the performance of ViT-UNets on segmenting smaller organs (e.g., Aorta and Pancreas). Furthermore, our novel multi-scale deformable positional encoding improved the DSC (\%) by 4.21\% for Gallbladder segmentation.

\section{Conclusion}
In this paper, we proposed AgileFormer that introduced spatially dynamic components (i.e., deformable patch embedding, spatially dynamic self-attention, and multi-scale deformable positional encoding) into a standard ViT-UNet for capturing spatially dynamic information for diverse target objects in the medical image segmentation. Extensive experiments demonstrated the effectiveness of the proposed method for a variety of medical image segmentation tasks. In addition, our AgileFormer significantly outperformed recent  2D and 3D segmentation methods, setting a new state-of-the-art performance.
We aspire that the idea of systematically introducing spatially dynamic components to ViT-UNet will guide future designs on how to extract spatially dynamic representations for medical image segmentation with multiple targets, which involve heterogeneous appearances. 

{\small
\bibliographystyle{ieee_fullname}
\bibliography{egbib}
}

\appendix
\section*{Appendix}

\renewcommand{\thefigure}{S\arabic{figure}}
\setcounter{figure}{0}
\renewcommand{\thetable}{S\arabic{table}}
\setcounter{table}{0}

\begin{table*}[!t]
    \centering
      \caption{Optimal training hyperparameter and model configurations for different experimental settings.}
      \resizebox{0.99\textwidth}{!}{
    \begin{tabular}{ll|c|c|c|c|c|c|c}
         \toprule
          & & \multicolumn{4}{c|}{\textbf{ training hyperparameters}} & \multicolumn{3}{c}{\textbf{model configuratoins}} \\
          \cmidrule(r){3-6} \cmidrule(r){7-9} 
          & \textbf{Models} & learning rate & epochs & batch size & \makecell{weight balance \\ parameter $\lambda$} & \makecell{first  deformable\\ patch embedding \\ strides} & \makecell{downsampling \\ patch embedding  \\ strides} & \makecell{input size} \\
        \toprule
         \multirow{2}{*}{\hfil \textbf{2D}} & AgileFormer-B [Synapse multi-organ] & 2e-4 & 400 & 24 & 0.6 & [2, 2], [2, 2] &  \makecell{[2, 2], [2, 2], [2, 2]} & $224 \times 224$  \\
        & AgileFormer-B [ACDC cardiac] & 4e-4 & 400 & 24 & 0.6 &  [2, 2], [2, 2] &  \makecell{[2, 2], [2, 2], [2, 2]} & $224 \times 224$  \\
        \midrule
        \multirow{3}{*}{\hfil \textbf{3D}} &  AgileFormer-T [Synapse multi-organ] & 2e-4 & 1000 & 4 & 0.5 &  [1, 2, 2], [2, 2, 2] &  \makecell{[2, 2, 2], [2, 2, 2], [2, 2, 2]} & $64 \times 128 \times 128$  \\
        & AgileFormer-T [ACDC cardiac] & 3e-4 & 1000 & 4 & 0.5 &  [1, 2, 2], [2, 2, 2] &  \makecell{[1, 2, 2], [1, 2, 2], [2, 2, 2]} & $14 \times 160 \times 160$ \\ 
        & AgileFormer-T [Decathlon brain tumor] & 5e-4 & 200 & 4 & 0.5 &  [2, 2, 2], [2, 2, 2] &  \makecell{[2, 2, 2], [2, 2, 2], [2, 2, 2]} & $96 \times 96 \times 96$ \\ 
        \bottomrule 
    \end{tabular}}
    \label{tab:hyperparams}
\end{table*}

\section{Comparison to CoTr~\cite{xie2021cotr} and MERIT~\cite{rahman2024multi}}
\begin{figure}[ht]
    \centering
    \includegraphics[width=1.0\columnwidth]{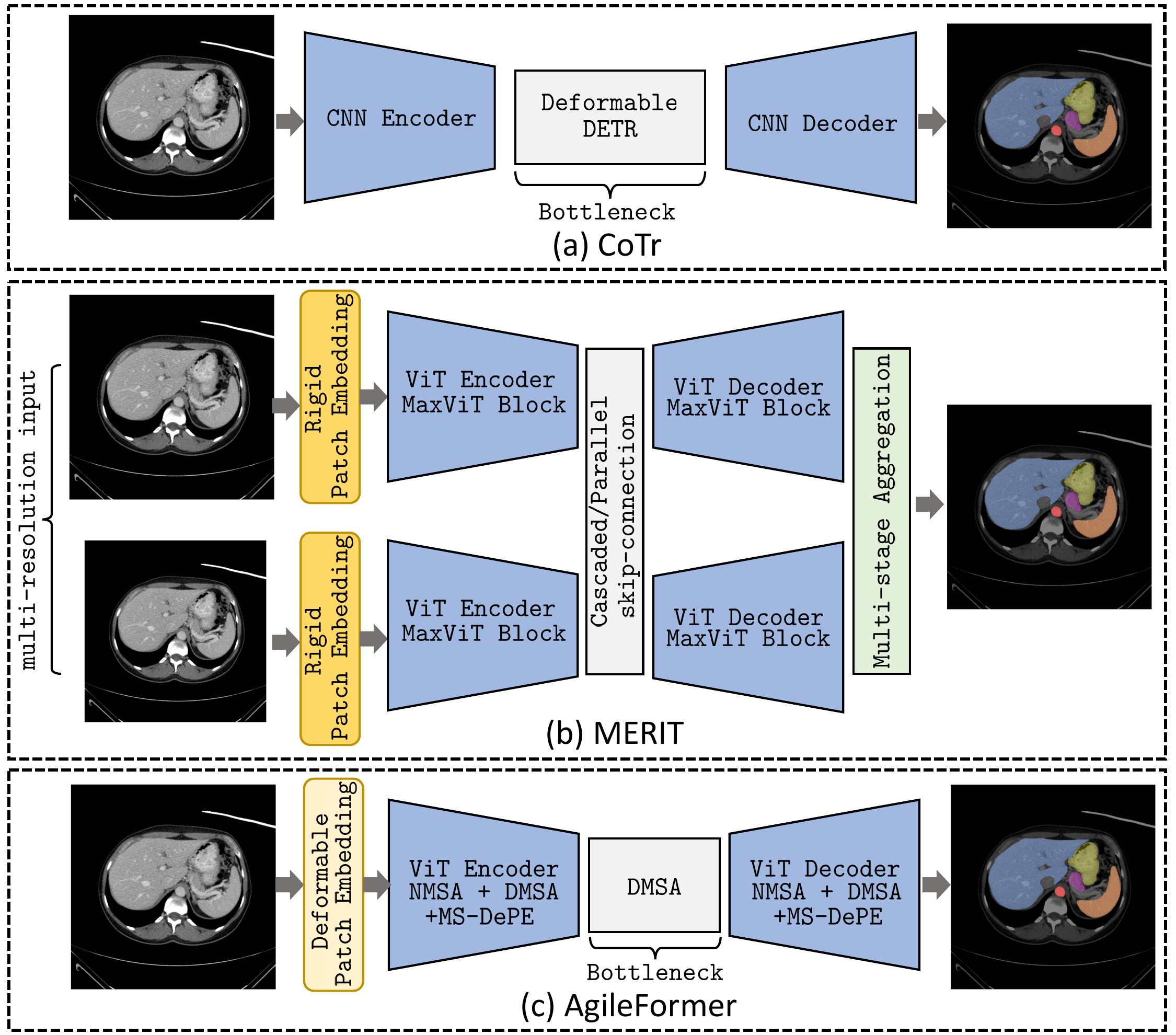}
    \caption{(a) CoTr is a hybrid model where Deformable DETER~\cite{zhu2020deformable} is only applied to the bottleneck. (b) MERIT captures dynamic feature representations using a multi-resolution and multi-stage approach, with the MaxViT~\cite{tu2022maxvit} as the building block. (c) AgileFormer alternates NMSA and DMSA as the building block, adding novel deformable patch embedding and multi-scale deformable positional encoding.}
    \label{fig:appendix:comparison}
\end{figure}
\begin{table*}[!t]
    \centering
    \caption{Ablation studies on model design variants in Synapse multi-organ segmentation task, starting from a \textcolor{darkgray}{SwinUNet baseline}. This is also the roadmap going from SwinUNet to the proposed AgileFormer.}
    \resizebox{0.99\textwidth}{!}{
    \begin{tabular}{c|c|cccccccc|c}
    \toprule
         & Avg DSC$\uparrow$ & Aorta & Gallbladder & Kidney(L) & Kidney(R) & Liver & Pancreas & Spleen & Stomach & $\substack{\text{Parms (M)} \\ / \text{FLOPS (G)}}$ \\
         \rowcolor{darkgray!8}
        \textcolor{darkgray}{Baseline (WMSA + RPE)} & \textcolor{darkgray}{79.13}  & \textcolor{darkgray}{85.47} & \textcolor{darkgray}{66.53} & \textcolor{darkgray}{83.28} & \textcolor{darkgray}{79.61} & \textcolor{darkgray}{94.29} & \textcolor{darkgray}{56.58} & \textcolor{darkgray}{90.66} & \textcolor{darkgray}{76.60} & \textcolor{darkgray}{27.17 / 6.13}  \\
        \hline
         Convolutional Downsampling & 79.40 & 85.74 & 70.56 & 82.33 & 80.69 & 93.50 & 58.89 & 91.46 & 72.05 & 29.10 / 6.34  \\ 
         Deformable Patch Embedding & 80.91 & 87.38 & 71.25 & 85.17 & 83.08 & 94.11 & 58.71 & 88.92 & 78.67 & 29.14 / 6.54  \\ 
        \hline
         NMSA + WMSA (w/o PE) & 81.27 & 87.22 & 68.79 & 84.73 & 79.49 & 95.31 & 63.13 & 90.62 & 80.86 & 26.74 / 6.31  \\ 
         WMSA + DMSA (w/o PE) & 81.47 & 86.12 & 67.37 & 84.18 & 80.13 & 95.61 & 64.86 & 90.97 & 82.55 & 26.77 / 5.98  \\ 
         NMSA + DMSA (w/o PE) & 82.06 & 87.54 & 70.95 & 81.90 & 79.29 & 95.36 & 68.03 & 90.72 & 82.72 & 26.82 / 5.99  \\
        \hline
         + APE & 81.96 & 88.35 & 70.72 & 81.82 & 78.84 & 95.09 & 67.80 & 90.27 & 82.45 & 28.90 / 5.99 \\ 
         APE $\rightarrow$ deformable RPE~\cite{xia2022vision} & 82.42 & 87.17 & 73.32 & 86.08 & 82.54 & 95.08 & 65.65 & 88.64 & 80.92 & 26.99 / 6.00 \\ 
        deformable RPE $\rightarrow$ CPE & 81.91 &  88.13 & 72.28 & 79.81 & 77.75 & 95.06 & 65.24 & 91.90 & 85.14 & 26.87 / 6.01 \\ 
        CPE $\rightarrow$ MS-DePE & 82.74  & 88.08 & 75.16 & 82.41 & 81.36 & 95.09 & 67.23  & 90.94 & 81.61 & 27.47 / 7.08\\ 
        \hline
         + Deep Supervision & 83.59 & 88.81 & 74.43 & 84.61 & 82.78 & 95.48 & 69.45 & 90.14 & 83.05 & 28.85 / 7.36\\ 
    \bottomrule
    \end{tabular}}
    \label{tab:ablation}
\end{table*}

\subsection{Comparison to CoTr~\cite{xie2021cotr}}
Similar to its ancestor model~\cite{zhu2020deformable}, CoTr is a hybrid model where the main backbone (encoder/decoder) is still a CNN. Multi-scale deformable attention outlined in~\cite{zhu2020deformable} serves as a bottleneck (see~\Cref{fig:appendix:comparison} (a)). As a result, it can only introduce limited deformability to capture spatially dynamic feature representations, as the main feature extractor does not have spatially dynamic components. In addition, CoTr does not take full advantage of ViT to capture long-range dependencies as its main backbone is still a CNN. We hypothesized that these factors may limit its performance and scalability. Instead, the proposed method uses a pure ViT as the main backbone (encoder/decoder), where neighborhood attention and deformable attention~\cite{xia2022vision} serve as the main building blocks (see~\Cref{fig:appendix:comparison} (c)). This better preserves the advantage of ViT while having the additional advantage of capturing spatially localized and dynamic features. This accounts for the superior performance of AgileFormer compared to CoTr. In particular, AgileFormer outperforms CoTr by 5.03,  1.24, and 1.6 in DSC (\%) for multi-organ, cardiac, and brain tumor segmentations, respectively. 

\subsection{Comparison to MERIT~\cite{rahman2024multi}}
MERIT mainly takes advantage of multi-resolution inputs to capture representation at multiple scales (see~\Cref{fig:appendix:comparison} (b)). This is achieved by combining the features extracted from inputs of different sizes. As a result, MERIT mainly models targets with varying sizes but may fail to capture objects with varying shapes. 
Instead, our AgileFormer handles both varying sizes as well as varying shapes by deformable patch embedding and spatially dynamic attention (see~\Cref{fig:appendix:comparison} (c)). In addition, we proposed a multi-scale deformable positional encoding to further capture multi-scale information. Consequently, AgileFormer outperforms MERIT by 0.84 and 0.23 in DSC (\%) for multi-organ and cardiac segmentation, respectively. 

\noindent \textbf{More comparisons:} We would like to point out that the official implementation~\footnote{\url{https://github.com/SLDGroup/MERIT}} of MERIT used test set to select the best model on Synapse. 
In contrast, we did \underline{not} use the test set for model selection and only evaluated the model checkpoint at the last training epoch. As stated in Table 1, if using the test set to select the best model, Agileformer can achieve an average DSC (\%) of \underline{86.11 (vs. 84.90} MERIT). To be more rigorous, if evaluated on the last training checkpoint, MERIT only achieved a DSC (\%) of \underline{82.89 (vs. 85.74} AgileFormer) on Synapse. Therefore, the improvement over MERIT is not marginal but statistically significant.  

\section{Additional Implementation Details}

The model configurations and optimal training hyperparameter settings for 2D and 3D experiments are shown in~\Cref{tab:hyperparams}. We used different downsampling strides for patch embeddings to accommodate different input sizes. The weight parameters $\lambda$ balance the DSC loss ($\mathcal{L}_{\text{DSC}}$) and cross-entropy loss ($\mathcal{L}_{\text{cross-entropy}}$)  as follows: 
\begin{equation}
    \mathcal{L} = \lambda \mathcal{L}_{\text{DSC}} + (1 - \lambda) \mathcal{L}_{\text{cross-entropy}}.
\end{equation}


\section{Ablation studies}
Here, we provide detailed results for ablation studies in~\Cref{tab:ablation}, including the change of segmentation accuracy, parameters, and floating points (FLOPs).

\section{Additional Qualitative Results}
\begin{figure*}[!t]
    \centering
    \includegraphics[width=1.0\textwidth]{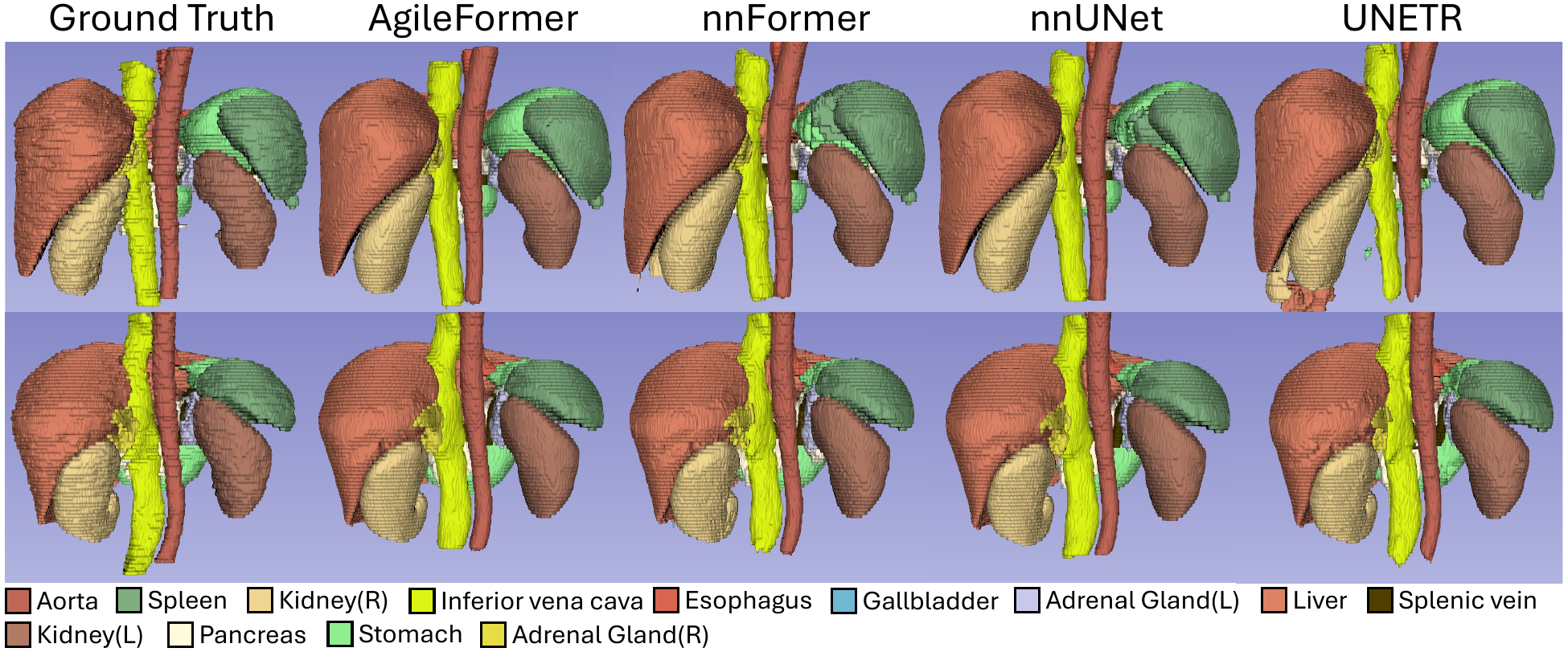}
    \caption{ Qualitative comparison between the proposed AgileFormer, nnFormer~\cite{zhou2021nnformer}, nnUNet~\cite{isensee2021nnu}, and UNETR~\cite{hatamizadeh2022unetr} on the Synapse multi-organ segmentation task.}
    \label{fig:vis}
\end{figure*}

\begin{figure*}[!t]
    \centering
    \includegraphics[width=1.0\textwidth]{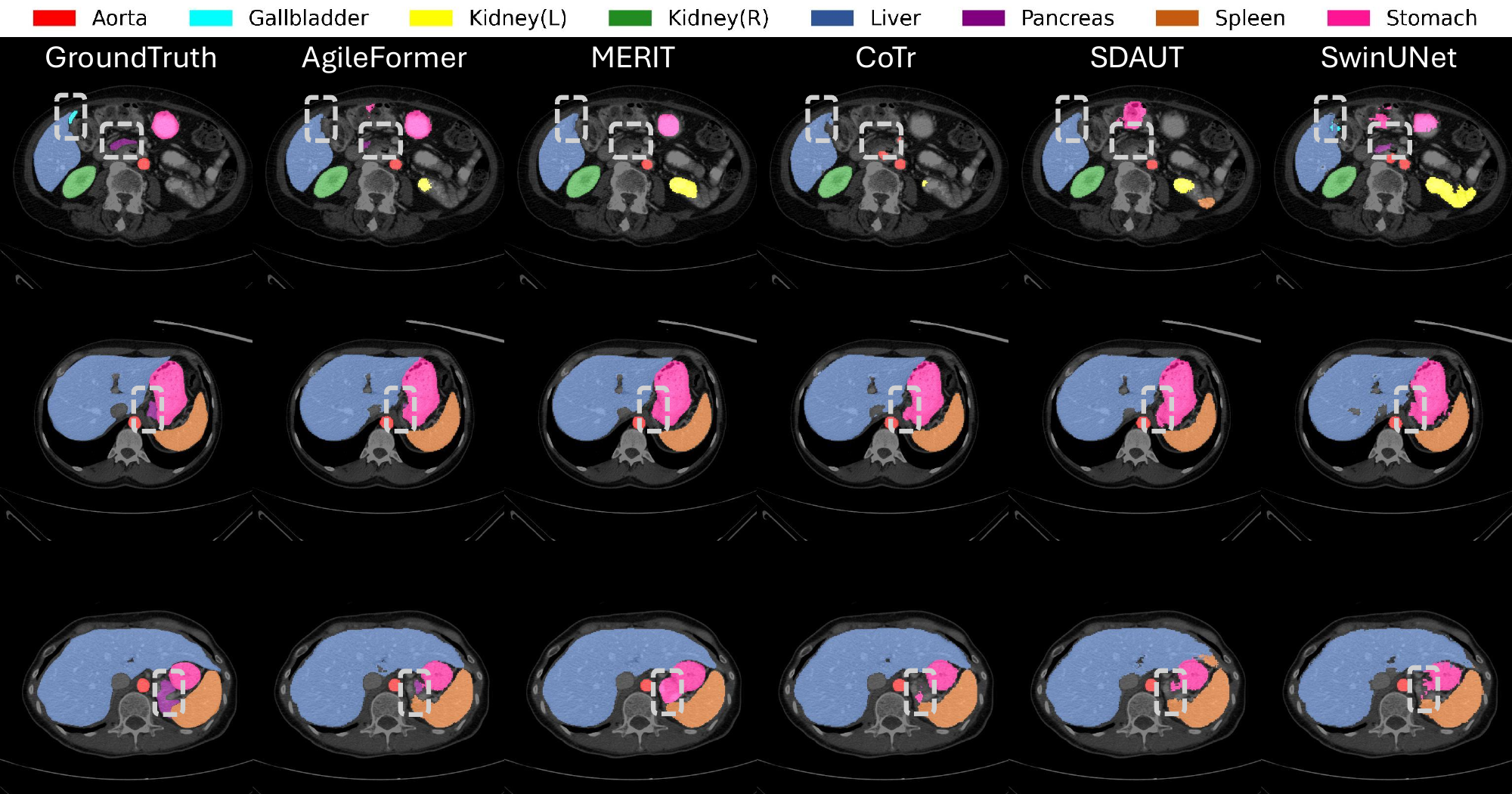}
    \caption{Failure cases where accurately segmenting irregular organs like the gallbladder and pancreas (highlighted by \textcolor{gray}{gray boxes}) is challenging for most methods with spatially dynamic components~\cite{huang2022swin,rahman2024multi,xie2021cotr}, including AgileFormer.}
    \label{fig:failure_synapse}
\end{figure*}

\begin{figure*}[!t]
    \centering
    \includegraphics[width=1.0\textwidth]{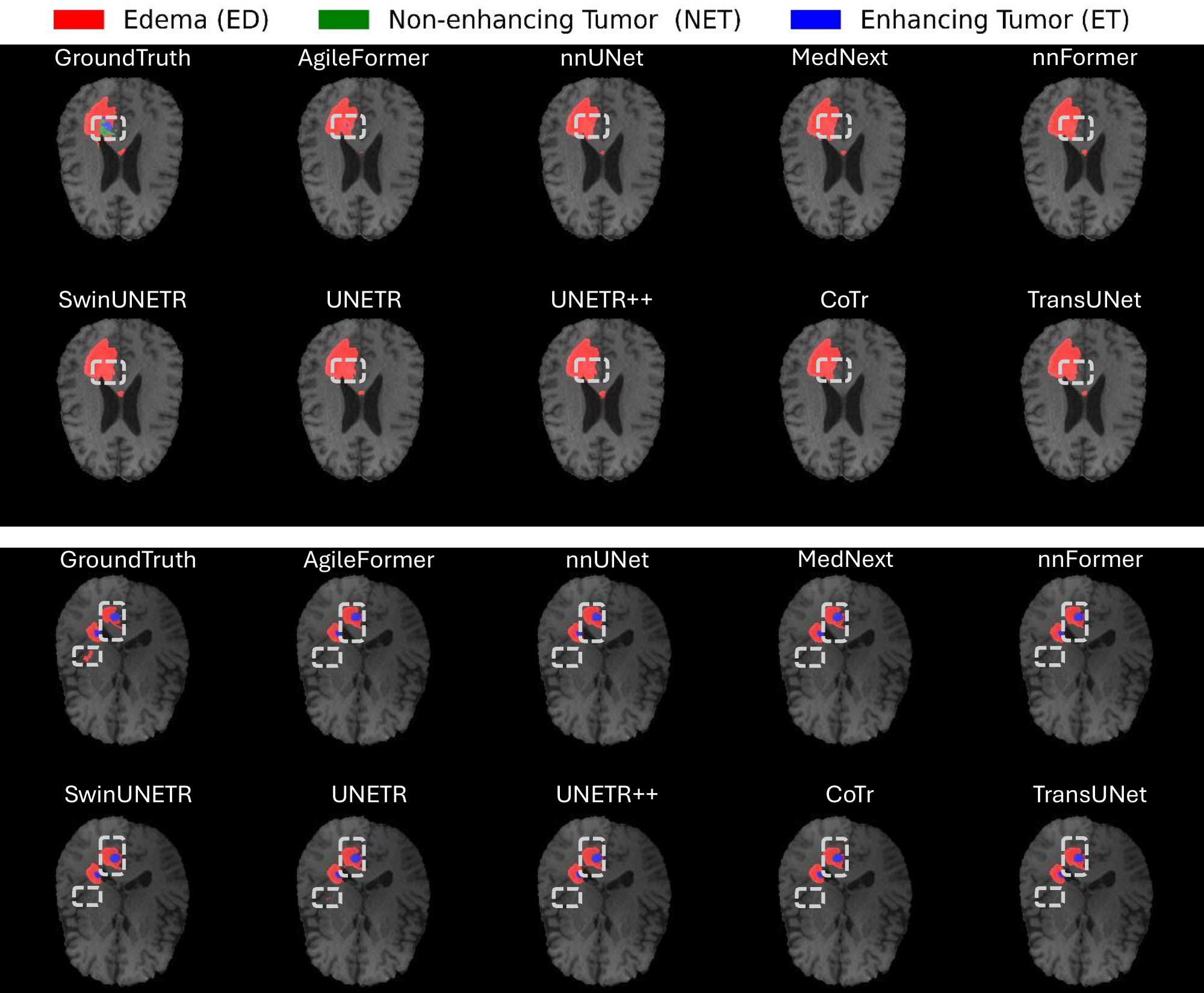}
    \caption{Failure cases where accurately segmenting irregular tumor tissues (highlighted by \textcolor{gray}{gray boxes}) is challenging for most methods~\cite{isensee2021nnu,roy2023mednext,chen2021transunet,xie2021cotr,hatamizadeh2022unetr,shaker2024unetr++,hatamizadeh2021swin,zhou2021nnformer}, including AgileFormer.}
    \label{fig:failure_tumor}
\end{figure*}

Additional qualitative results on Synapse produced by 3D models are shown in~\Cref{fig:vis}, where AgileFormer showcased accurate segmentation compared to other state-of-the-art methods. Some failure cases on Synapse are shown in~\Cref{fig:failure_synapse}, where even methods that can extract spatially dynamic features (including AgileFormer) cannot accurately segment highly irregular organs such as Gallbladder and Pancreas. Similarly,~\Cref{fig:failure_tumor} shows failure cases on brain tumor segmentation tasks, where tissue classes with highly irregular appearance cannot be accurately segmented by most methods, including AgileFormer.

\section{Statistical Test}
On Synapse, we performed the Wilcoxon signed rank test to the second-best model (i.e., 2D MERIT and 3D nnUNet), where the populations are the DSC values for 8 different organs. However, on ACDC and Decathlon datasets, we performed the paired t-test to the second-best model, as there are only three foreground labels. In these two cases, the populations were the DSC values for all subjects across the dataset. 

\end{document}